\documentclass[10pt,twocolumn,letterpaper]{article}

\usepackage{iccv}
\usepackage{times}
\usepackage{epsfig}
\usepackage{graphicx}
\usepackage{amsmath}
\usepackage{amssymb}
\usepackage{ulem}
\usepackage{multirow}

\usepackage[breaklinks=true,bookmarks=false]{hyperref}

\iccvfinalcopy 

\def\iccvPaperID{61} 

\ificcvfinal\fi

\begin{document}

\title{HealthWalk: Promoting Health and Mobility through Sensor-Based Rollator Walker Assistance}

\author{Ivanna Kramer, Kevin Weirauch, Sabine Bauer, Mark Oliver Mints, Peer Neubert\\
Institute for Computational Visualistics \\
University of Koblenz\\
{\tt\small \{ikramer, kweirauch, bauer, mmints, neubert\}@uni-koblenz.de}}

\maketitle
\ificcvfinal\thispagestyle{empty}\fi

\begin{abstract}
Rollator walkers allow people with physical limitations to increase their mobility and give them the confidence and independence to participate in society for longer. However, rollator walker users often have poor posture, leading to further health problems and, in the worst case, falls. 
Integrating sensors into rollator walker designs can help to address this problem and results in a platform that allows several other interesting use cases. This paper briefly overviews existing systems and the current research directions and challenges in this field. We also present our early HealthWalk rollator walker prototype for data collection with older people, rheumatism, multiple sclerosis and Parkinson patients, and individuals with visual impairments.
\end{abstract}

\section{Introduction}

\begin{figure}[h]
\centering
\includegraphics[width=1\linewidth]{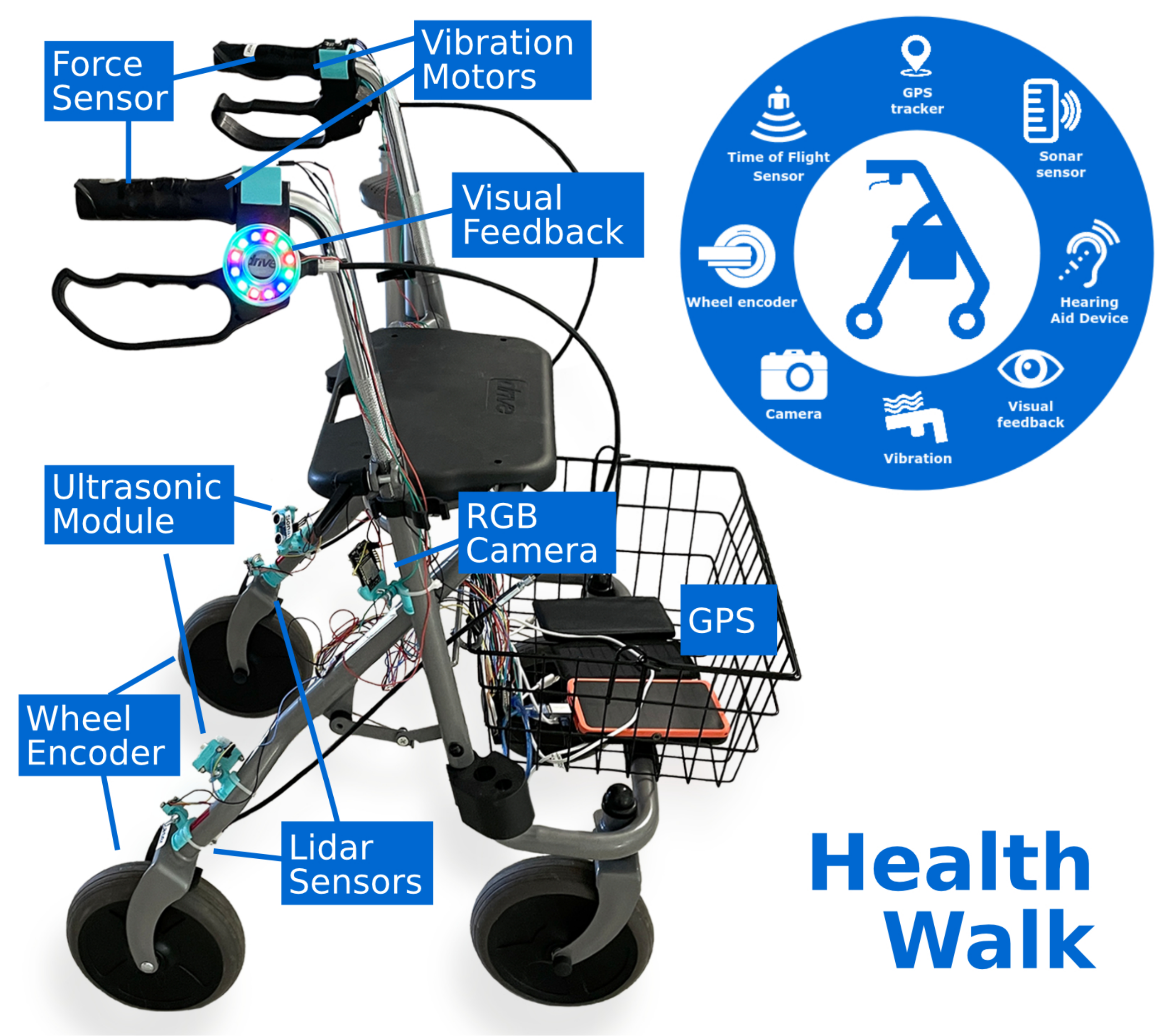}
\caption{Our early HealthWalk prototype version of a rollator walker includes force sensors on both handlebars, ultrasonic and lidar distance sensors for leg tracking and RGB cameras to capture the movement of user's lower body.} \label{fig:sensors_overview}
\vspace{-0.3cm}
\end{figure}


Many elderly people and people with physical limitations rely on walking aids. A widely used walking aid is the rollator walker, which helps people with physical limitations get around independently while providing a measure of safety. However, the lack of individualized ergonomic adjustment options and aids, as well as posture guidance and feedback systems, leads to certain risks when using the rollator walker. This often results in the user unconsciously adopting an non-ergonomic posture by bending their upper body forward. Furthermore, especially for users who are constantly dependent on rollator walkers, a flexed posture may contribute to the chronicity of back and shoulder pain, tendonitis, carpal tunnel syndrome, and osteoarthritis \cite{Bateni2005ADF}. Similarly, it appears that flexed posture and improper use of a rollator walker lead to an increased risk of falls \cite{Liu2009AOR}. These falls are often accompanied by fractures that require surgical treatment and that have a dangerously high post-surgical mortality, for example, after fractures of the neck of the femur \cite{Bzovsky2020FAW,Ge2023ABH}. 

A promising direction to address the aforementioned problems is to extend the existing predominantly purely mechanical rollator walkers with sensing, information processing, and user-interaction capabilities. Moreover, such sophisticated rollator walkers provide a platform for further assistance systems like training and rehabilitation programs (e.g., in combination with gamification aspects), monitoring of user mobility, assistance for localization and navigation, data collection for city infrastructure development, and many more. 

\begin{table*}[t]
\centering
\small\addtolength{\tabcolsep}{0pt}
\begin{tabular}{|l| c| c| } 
\hline
\textbf{Application} & \textbf{Onboard sensors}  &  \textbf{References} \\ 
\hline \hline
\multirow{2}{*}{Fall risk assessment}
    & Force sensor in handlebars, wheel encoders, RGB-D camera,  
    & \multirow{2}{*}{\cite{ballesteros2018automatic,merlet2012ang}} \\    
    & 3D accelerometer, gyrometer  infra-red range sensors,  GPS, RGB camera  &\\
\hline
\multirow{4}{*}{\shortstack[l]{Obstacle detection\\ Navigation}}
    &   RGB-D camera, time of flight (ToF) distance sensors,                    
    &  \multirow{4}{*}{\cite{feltner2019smart,fernandez2022walk,grzeskowiak2022swalkit,macnamara2000smart,miro2009robotic,sierra2021assessment}} \\          
    &  inertial measuremet unit (IMU), ultrasonic sonar distance sensors,    &   \\
    &   wheel encoders, 2D lidar,   triaxial load cells,   &   \\
    &   2D laser rangefinder, linear hall-effect sensor     &   \\

\hline
\multirow{3}{*}{Gait analysis}
    &   Force sensors, doppler radar sensor, 3D accelerometer,    
    &  \multirow{3}{*}{ \cite{fernandez2022walk,postolache2011smart,valadao2014towards}} \\ &  triaxial force sensor, laser sensor,    &   \\  
    &   wheel encoder, strain gauges, lidar (360° FoV)  &   \\

\hline
\multirow{3}{*}{\shortstack[l]{Physical condition\\ monitoring  }}
    & Wheels with magnets  binary pressure sensor in seat,
    & \multirow{3}{*}{\cite{chan2008smart}} \\   
    & strain gauge, hall-effect sensor,  tri-axial accelerometer,  &   \\
    & sensors for cardiac and blood oxygenation monitoring  &   \\

\hline
\end{tabular}
\caption{Summary of sensors incorporated in the sensor-based rollator walker aids (grouped by application).}
\label{tab:sensors_literature}
\vspace{-0.3cm}
\end{table*}

In this paper, we first provide a short overview of existing sensor-based rollator walker aids together with their sensor setup and grouped by applications in Section~\ref{sec:sota}.
Section~\ref{sec:research_questions} identifies and outlines the interdisciplinary research questions that arise in this context.
Finally, Section~\ref{sec:project} introduces our HealthWalk rollator walker prototype depicted in Figure~\ref{fig:sensors_overview} with modular sensor and application concepts and presents our plans to use it for data collection.

\section{Existing Sensor-Based Rollator Walker Aids}
\label{sec:sota}

There are multiple rollator walkers designed for the mobility assistance of  a specific user population in certain tasks, such as fall detection or navigation, by using various sensors and actors attached to the rollator walker. Dupret at al. \cite{dupret2022erste} outlines the high interest of the rollator users to apply sensory-based assistance, however the sensors have to be adapted to the varying requirements of the specific target groups. 

In recent years, numerous sensor-based mobility rollator walkers have been developed to assist users with specific needs in various scenarios. Applications include 
fall risk assessment \cite{ballesteros2018automatic,merlet2012ang}, 
Obstacle detection and navigation \cite{feltner2019smart,fernandez2022walk,grzeskowiak2022swalkit,macnamara2000smart,miro2009robotic,sierra2021assessment},
gait analysis \cite{fernandez2022walk,postolache2011smart,valadao2014towards}, and
physical condition monitoring \cite{chan2008smart}.
Table \ref{tab:sensors_literature} presents an overview of the used sensors for these tasks. RGB-D cameras and proximity sensors, are used for fall detection as well as for obstacle detection and navigation. In addition, force sensors in the handlebars and lidars on the walker frame are used primarily for gait analysis.

\section{Research Directions and Challenges}
\label{sec:research_questions}

The development of a sensor-based walker, tailored to the needs of a specific target group and the challenges of long-term use, creates a diverse research landscape involving multible disciplines. This multidisciplinary space includes research perspectives from mechanical and electrical engineering, signal processing, computer science, healthcare and human-machine interaction studies. 

The mechanical design, construction, and materials are essential aspects of any assistive device.
The general requirements for extending these mechanical systems with perception and interaction capabilities include affordability, small size and low weight, dependability and robustness, diversity of sensing modalities, sensing range and accuracy, safety, energy consumption, and low computational demands.
The expected extensive collection of sensor data in private and public domains raises high demands for data privacy.
A particular challenge is the integration into or combination with existing rollator walker designs.

\subsection{Human-rollator walker interaction}
\label{sec:challenge_interaction}

A well-designed user-centric interaction system for rollator walkers can enhance usability and positively impact the user's physiological and cognitive condition, building trust in the device. The challenge lies in catering to different target groups with various physical and non-physical restrictions, necessitating different types of interactions. While the visual feedback would be conceivable for the users with Parkinson decease to initiate the walking after gait freeze, individuals with visual limitations  would find audio prompts more advantageous, providing essential assistance during navigation. Most of the presented rollator walkers have implemented a feedback system with haptic \cite{feltner2019smart, grzeskowiak2022swalkit} or audio signals \cite{MacNamaraASWF2000}. However, more research is needed to investigate which kind of feedback would be suitable or preferable for specific target group and adaptable to the environmental changes.

\subsection{User perception}
\label{sec:challenge_user_perception}
We can distinguish two directions regarding sensor perception: The rollator walker can observe the environment (see Section ~\ref{sec:challenge_env_perception}) and it can observe the user.
For the latter, a general research question is how to assess the physiological state of the user from sensory data that provide only a partial view of the user's body and do not capture internal physiological parameters.

During rollator walker use the user tends to adopt non-physiological body postures for various reasons \cite{liu2009assessment}. These include improper setup of the rollator walker to match the user's size and weight or the presence of certain medical conditions or neurological disorders. Yuanyuan et al. \cite{guo2020postural} highlighted that the unnatural forward-leaning posture adopted while using a walker significantly affects gait patterns in female seniors. Maintaining an upright posture is essential for efficient mobility and reducing the risk of falls. 
Therefore body posture and user condition assessment from limited data are important research questions.
 
Multisensoric gait and posture analysis approaches can include distance modules, force sensor response from handlebars, and video recordings explicitly focusing on the user's lower extremities. 
A challenge arises from the limited field of view covered by sensors, as due to ethical and privacy reasons, vision sensors might not capture the upper body parts (from hip upwards). Additionally, the individual nature of the user's posture and gait makes generalizing posture characteristics from gait analysis a difficult but captivating task. Promising approaches might include biomechanical models and simulations of the user's motion. 


Another important goal is the prevention or at least detection of falls. Falls can be caused by external reasons, like bumping into an obstacle, but the users' balance can get thrown off for other reasons as well. Multi-variate time series analysis of the sensor measurements can be used to 1) predict potential falls as early as possible and 2) if falls happened detect them accurately and reliably. 
Concerning the human-rollator interaction (cf. Section~\ref{sec:challenge_interaction}), the first case raises the additional challenge of identifying appropriate reactions to prevent the fall, e.g., alarming feedback to the user or activating automated braking. In the second case of an actual fall, an integrated emergency call device in the rollator walker might be activated. However, this poses high demands on the reliability of distinguishing a fall from any other uncritical movement of the rollator walker.

\subsection{Environmental perception}
\label{sec:challenge_env_perception}
When navigating the environment using a rollator walker, some important challenges come up. For example, users have difficulty moving from low to high ground and vice versa. Namely, this comes up in the context of staircases and sidewalks. Furthermore, the surface characteristics of the walking area can introduce additional restrictions to the user's mobility.
Therefore we assume that environmental perception is key for people to be safe when using a rollator walker. Established techniques from the field of (Visual) SLAM (simulations localization and mapping), terrain classification, or path planning from the field of mobile robotics can be used. Particular challenges arise due to the limitations on the price, size, weight, and mounting positions of sensors and the available energy and compute resources.

In addition to static obstacles, there are also dynamic hazards. A person using a rollator walker in a pedestrian precinct or a busy area needs to be able to perceive the environment quickly and correctly, for example when crossing a street. Even more so if their own perception is limited, for example by a visual impairment.

\subsection{Use-cases with additional stakeholders}
\label{sec:challenge_other_stakeholders}
The rollator walker, acting as a mobile sensor carrier, can be an important source of information for various stakeholders. For example, caregivers and relatives can receive real-time updates on the user's location and status. Doctors and physiotherapists can monitor and analyze the user's activity levels and characteristics. Additionally, combining the rollator's environmental perception with user motion patterns can offer valuable insights for designing barrier-free buildings and infrastructure, highlighting areas that require improvement. However, these applications raise significant data privacy concerns and pose challenging research questions.


\section{The HealthWalk Prototype}
\label{sec:project}

Our goal is to provide personalized assistance and body support to older adults and individuals with physical limitations by introducing sets of sensors that can be incorporated into the design of different existing rollator walkers in a modular manner to address specific populations' needs. The modularity of the integrated sensors allows for customization and adaptability, promoting varying levels of mobility and support. These modular sensor kits can seamlessly integrate into a conventional rollator walker, ensuring a user-friendly experience.
Further, we aim to develop algorithmic solutions that use the sensor data collected by the rollator walker to help users maintain their physiological posture and safety during mobility by overcoming the challenges that arise due to environmental and personal restrictions. 
We want to address the environmental perception based on Visual SLAM (cf. \cite{gao2021introduction}). For time series analysis, body posture and physiological condition assessment we plan to use high dimensional computing \cite{schlegel2022hdcminirocket}. To develop a suitable rollator feedback system for diverse target groups, our approach involves actively engaging the affected individuals (active participation method) to collect and analyze the requirements for establishing an effective human-rollator walker interaction.

The current version of the HealthWalk rollator walker, illustrated in  Figure~\ref{fig:sensors_overview}, is designed to help users maintain a healthy upright posture while walking. This walking aid is equipped with two SEN-PRESSURE10 pressure sensors on the handlebars, two VL53L0X ToF distance sensors, and two HC-SR04 sonar modules attached to the frame. Additionally, four ESP32 RGB cameras capture the movement of the user's body from the hip downwards. These sensory data can be used to detect and monitor various physiological and physical parameters of the rollator walker user (\textit{user perception}). Finally, a GPS tracker is integrated into the rollator walker to enable location tracking. Extra proximity sensors and an RGB-D camera at the rollator walker's frame can be used to collect and analyze the information about the surrounding environment of the rollator walker and user (\textit{environmental perception}). 

In the upcoming project phase, we plan to collect a comprehensive dataset capturing motion data using a rollator walker by older people, rheumatism, multiple sclerosis, Parkinson's patients, and individuals with visual impairments. Thus, we aim for the dataset to cover the user's lower body, including gait and the surrounding environment, from different data modalities captured by the proposed sensors. The protocol and data collection requirements would be a part of a discussion in the ongoing workshop.





\section{Conclusion}

We see sensor-based rollator walkers as a promising approach to address problems like bad posture or falls and as an interesting platform for various further use cases. Moreover, the many practical and theoretical challenges make them also a very appealing research topic. The early HealthWalk prototype is intended to support research with collected datasets.


{\small
\bibliographystyle{ieee_fullname}
\bibliography{egbib}
}


\end{document}